\documentclass[11pt, a4paper, onecolumn, copyright, gdm]{google}

\usepackage[scaled]{helvet}
\usepackage[round]{natbib}
\usepackage{float}
\usepackage{fvextra}
\usepackage{tcolorbox}
\tcbuselibrary{skins,breakable}
\DeclareUnicodeCharacter{2026}{\ldots}
\DeclareUnicodeCharacter{2212}{-}

\definecolor{codebg}{HTML}{F8F9FC}
\definecolor{codeborder}{HTML}{E1E6EC}
\definecolor{captiontext}{HTML}{5F6368}
\renewcommand{\monthyeardate}{}\renewcommand{\today}{}
\newcommand{\renderedwithpack}{Rendered with the open-source Pixel Perfection texture pack (XSSheep and Nova\_Wostra, CC BY-SA 4.0), not Minecraft's textures.}
\newcommand{\iconsfrompack}{Item icons from the open-source Pixel Perfection texture pack (XSSheep and Nova\_Wostra, CC BY-SA 4.0), not Minecraft's textures.}
\newcommand{\iconsandfacesfrompack}{Item icons and player faces from the open-source Pixel Perfection texture pack (XSSheep and Nova\_Wostra, CC BY-SA 4.0), not Minecraft's textures.}
\uselogo{}
\title{A team of Gemini 4 Argon agents beats Minecraft}
\author[1]{Maxence Faldor}
\author[1]{Satinder Singh}
\author[1]{Roberta Raileanu}
\affil[1]{\thepa{}}

\begin{abstract}
We placed four Gemini 4 Argon agents in a sandboxed Minecraft survival world and gave them one goal: defeat the Ender Dragon. Each agent controlled a player via an open-source library called Mineflayer and coordinated with its teammates through in-game chat and shared memory. No one assigned roles or wrote game logic. The agents planned, wrote, and debugged every program themselves. Researchers could watch the run on a live dashboard and pause or stop it at any time, but they had no cause to intervene. Starting from a randomly generated world on Easy difficulty, the team wrote 2,783 programs and defeated the dragon in 27 hours and 14 minutes, dying 605 times along the way. To our knowledge, this is the first end-to-end autonomous completion of Minecraft by AI agents, achieved using only the standard Mineflayer library without pre-scouted seeds or other game-specific tools written by humans. Winning required the agents to keep a coherent plan across thousands of dependent steps for more than a day. Complex real-world software engineering and enterprise tasks demand the same long-horizon coding and multi-agent coordination. We describe the setup, how the run unfolded, the team's strategies (dividing the work, sharing what they learned, recovering each other's items), and the limits of the result.    
\end{abstract}

\begin{document}
\maketitle

\begin{figure}[H]
\centering
\includegraphics[width=\linewidth]{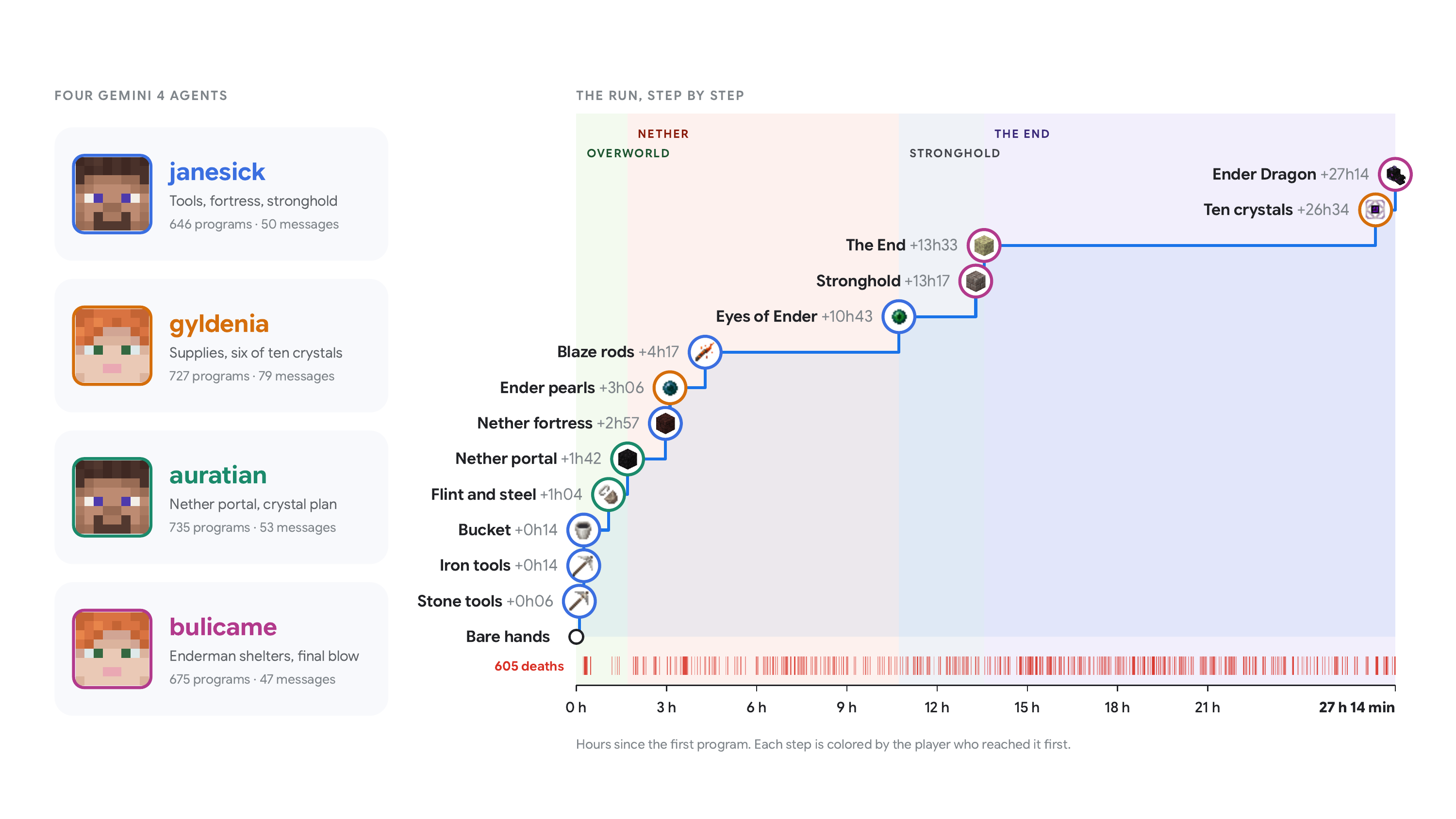}
\caption{The four agents, and their run step by step: the thirteen steps from bare hands to the Ender Dragon, each colored by the player who reached it first, over 27 hours and 14 minutes. \iconsandfacesfrompack}
\end{figure}

\section{Introduction}

Four Gemini 4 Argon agents entered a newly generated Minecraft survival world with one instruction: defeat the Ender Dragon. The run tests whether a team of general-purpose coding agents, not built for Minecraft, can pursue one goal for more than a day, through thousands of programs and hundreds of failures, while coordinating with its peers. Each agent controlled a player through Mineflayer, an open-source JavaScript library for Minecraft bots, and communicated with its teammates through the game's chat. Twenty-seven hours after their first program, the team defeated the dragon.

Reaching the end of Minecraft from a random spawn requires a long sequence of dependent steps across three dimensions (\citealp{fan2022minedojo}): gathering resources and crafting tools in the Overworld, collecting blaze rods and ender pearls in the Nether, locating an underground stronghold, and destroying the ten crystals that heal the dragon in the End. Prior work in Minecraft has trained policies on gameplay (\citealp{baker2022vpt}; \citealp{hafner2023dreamerv3}; \citealp{sima2_2025}) or paired language models with game-specific planners, curricula, and skill libraries (\citealp{wang2023describe}; \citealp{wang2024voyager}; \citealp{wang2025jarvis1}; \citealp{altera2024project}). Without pre-assigned roles or a prescribed plan, the four agents divided the work among themselves, kept shared notes of coordinates and lessons, recovered each other's items after deaths, and wrote 2,783 JavaScript programs to complete the game.

\begin{center}
\includegraphics[width=\linewidth]{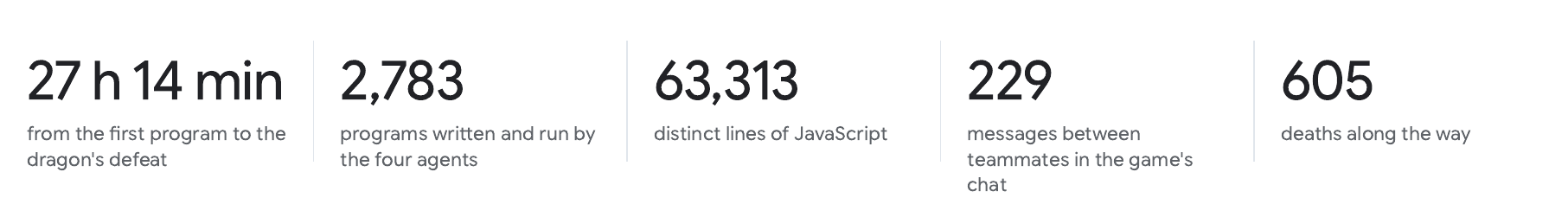}
\end{center}

\section{How the agents play}

Each player is driven by our general-purpose Gemini 4 Argon coding agent, connected to a standard Minecraft server through Mineflayer and its pathfinding plugin. Rather than seeing pixels or pressing keys, an agent acts by writing and running a short JavaScript program, such as \textit{walk to these coordinates and mine every iron ore within 16 blocks}, and by reading the text that comes back: its position, health and inventory, what it found, or the error it hit. It then writes the next program. Each program runs until its task finishes or reaches a time limit chosen by the agent; the median program ran for 33.5 seconds. Over the run, each player had a program running about 40\% of the time; the rest went to reading results, thinking and writing the next program.

Mineflayer lets a program read a player's position, health, inventory and surroundings, and order basic actions such as walking to a point, digging or placing a block, crafting an item or attacking. It does not know how to play: what to do, and in what order, was up to the agents.

\begin{figure}[H]
\centering
\includegraphics[width=\linewidth]{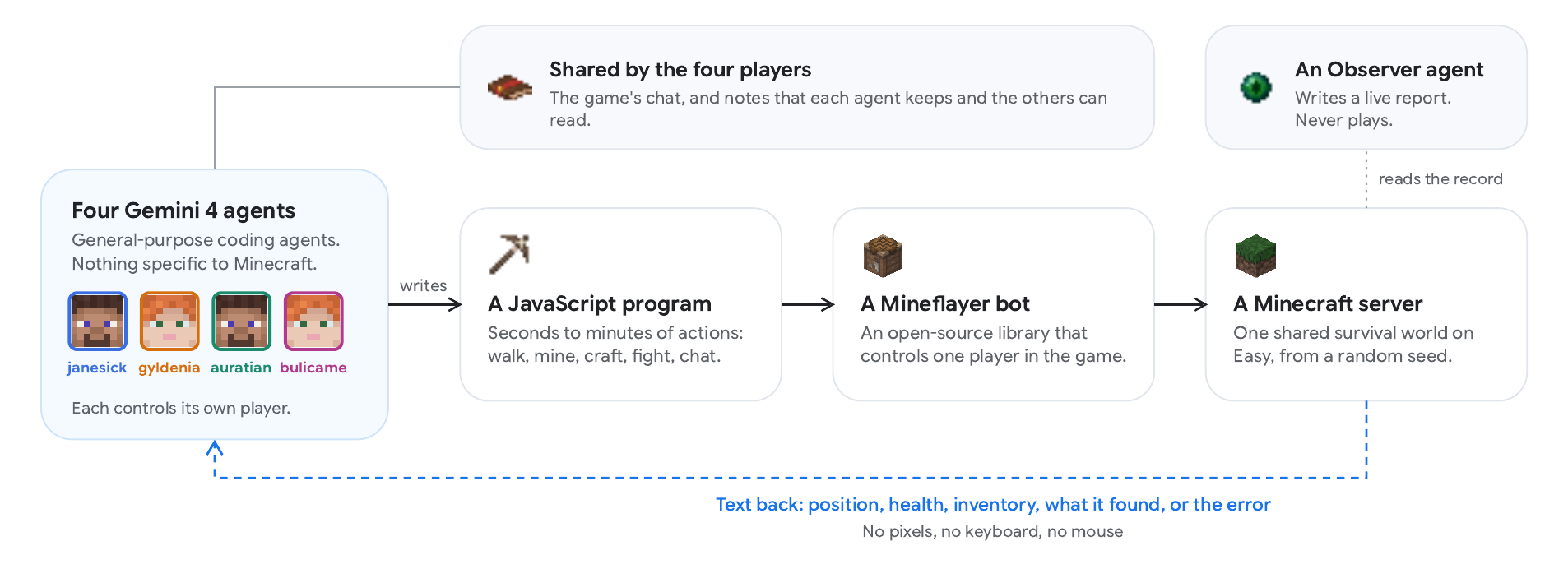}
\caption{How each agent plays. It writes a JavaScript program, a Mineflayer bot runs it in the shared world, and text comes back. The players share the game's chat and can read each other's notes. A fifth agent, the Observer, reports on the game but never plays. \iconsandfacesfrompack}
\end{figure}

The game, Minecraft Java Edition 1.20.4, ran in survival mode on Easy difficulty in a randomly generated world, with cheat commands disabled and no human input during the run. Beyond a brief note that they were working as a team of four, the agents' instructions about the game consisted of a single paragraph:

\begin{quote}
``The objective is to defeat the Ender Dragon, starting from the spawn of the session's shared Minecraft survival world. [\ldots{}] This is the only criterion. Any means of meeting it within the rules of survival mode is permitted, and the players decide how to share the work.''
\end{quote}

Their other instructions are the same as for any environment: verify the result of each action, build reusable tools, record lessons in a memory shared by the team, and report outcomes honestly.

\subsection{Agents and tools}

Each of the five agents, the four players and the Observer, is an instance of the same general-purpose coding agent, run with Gemini 4 Argon on its own cloud machine. Each has the tools it uses for software work, including a code workspace with version control, a shell, file editing and web search, and a persistent memory with a private part and a part shared by the team. The agent compacts its context when it grows too long, as it does for software work; the instructions ask agents to keep in memory what must survive a compaction, such as work in progress and next steps. None of the 2,839 programs the players ran during the session (2,783 before the win) made a network request or referred to a web source, and neither their chat nor the Observer's report mentions one.

Players interact with the game only through a command-line client. It sends a JavaScript program to the player's Mineflayer bot on the game server, waits up to a timeout that the agent chooses (60 seconds by default), and prints what the program logged and returned. The program runs as the body of an asynchronous function that receives the bot, with the pathfinding plugin loaded and no helpers added. It can load the built-in modules of Node.js and the installed Mineflayer packages, but cannot start processes or write files, and values it stores on the global object persist between programs. Players talk through the game's chat, which reaches every bot and appears in the output of each player's next program. The server records chat, deaths and advancements in its log, keeps every program with its output, and saves the world every five minutes; after a failure, it restarts from the latest save.

\subsection{Instructions}

Each agent's instructions combine a general part, the same for every environment, with a short description of the environment and of the objective. The general part for players asks them to observe the environment before acting and verify the result afterwards, to validate new tools on short tasks before trusting them, to build and refine their own tools when manual actions limit their speed or reliability, to record mechanics, strategies and workarounds in memory so that no obstacle is met twice, to keep working while background jobs run, to be honest about their actions and outcomes, and to pursue the objective rather than exploit flaws of the environment or of its evaluation. The description of the environment explains how to run code with the bot, how the chat works, where the server records the game and where to find the documentation of Mineflayer; it contains no game strategy. A brief note lists the other members of the team.

\subsection{Monitoring}

A fifth agent, the Observer, followed the session as an independent witness and never played. Its instructions ask it to check the players' claims against the record before reporting them, to report specification gaming, such as an exploited flaw in the evaluation, and fabricated results, to leave the game unchanged, and to intervene only on failures of the environment, such as a broken tool or a quota ceiling, never on the players' strategy. It published a live report of the session, with a summary of progress, a log of verified events and its findings about the infrastructure. It flagged no specification gaming, and it never joined the game or ran a program in it. The run didn't contain any safety incidents, according to the Observer agent and our manual reviews of the logs. 

\section{The 27-hour run}

From the first program to the dragon's defeat took 27 hours and 14 minutes. The four agents wrote and ran 2,783 programs, exchanged 229 messages and died 605 times along the way. That time includes 6 server restarts, which paused the world for about 20 minutes in all, and spans about 81 in-game days. For human players, HowLongToBeat's median for Minecraft's main story is 45 hours and the random-seed speedrun record is under seven minutes; the agents played as a team of four on Easy difficulty, through programs rather than a screen and keyboard.

\begin{figure}[H]
\centering
\includegraphics[width=\linewidth]{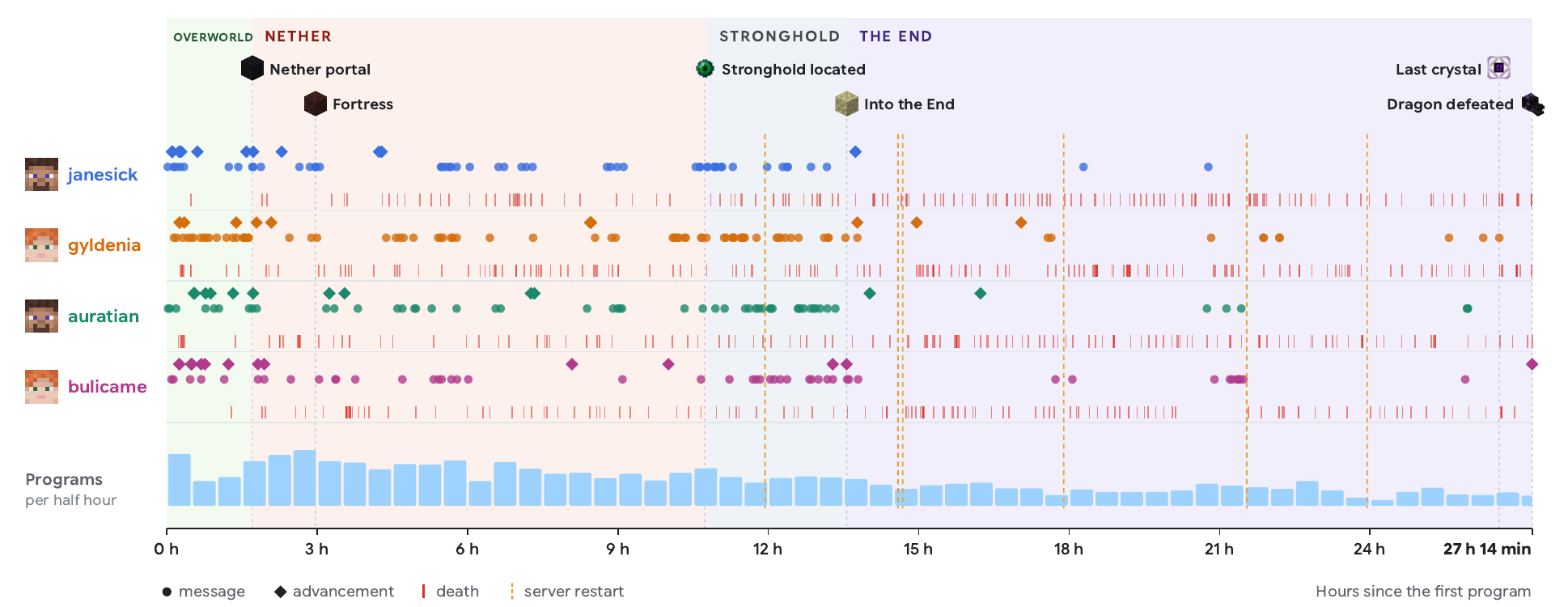}
\caption{The whole run. Each lane is a player: dots are chat messages, diamonds are advancements, red ticks are deaths. Bars count programs per half hour; orange dashes mark the six restarts of the world server. \iconsandfacesfrompack}
\end{figure}

\begin{figure}[H]
\centering
\includegraphics[width=\linewidth]{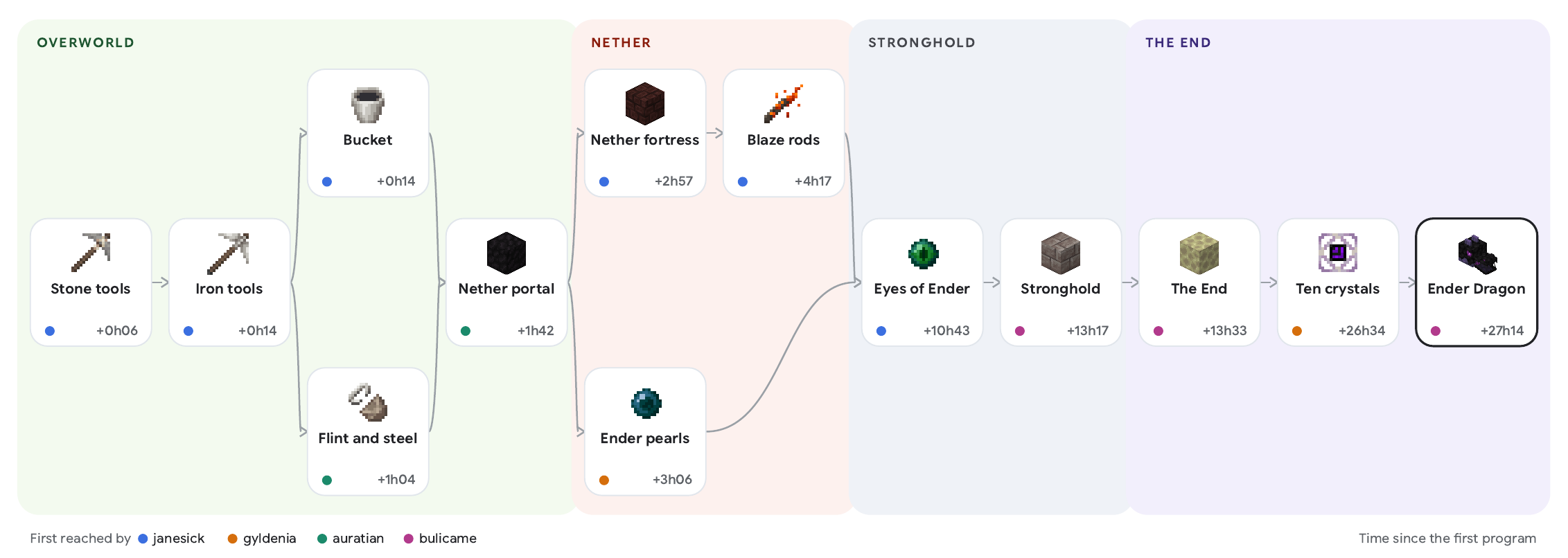}
\caption{The thirteen dependent steps from initial spawn to the Ender Dragon, each with the time the team first reached it and the player who did. \iconsfrompack}
\end{figure}

\subsection{Hour 0: a team forms}

Ten minutes after the first program ran, janesick proposed a division of labor in the chat. Within a minute after that, auratian answered with a different plan, and gyldenia replied ``Copy that!''. Roles settled quickly: gyldenia scouted for food, beds and flint; bulicame went down for diamonds; janesick smelted iron and made tools; auratian prepared the Nether portal.

\begin{figure}[H]
\centering
\includegraphics[width=0.78\linewidth]{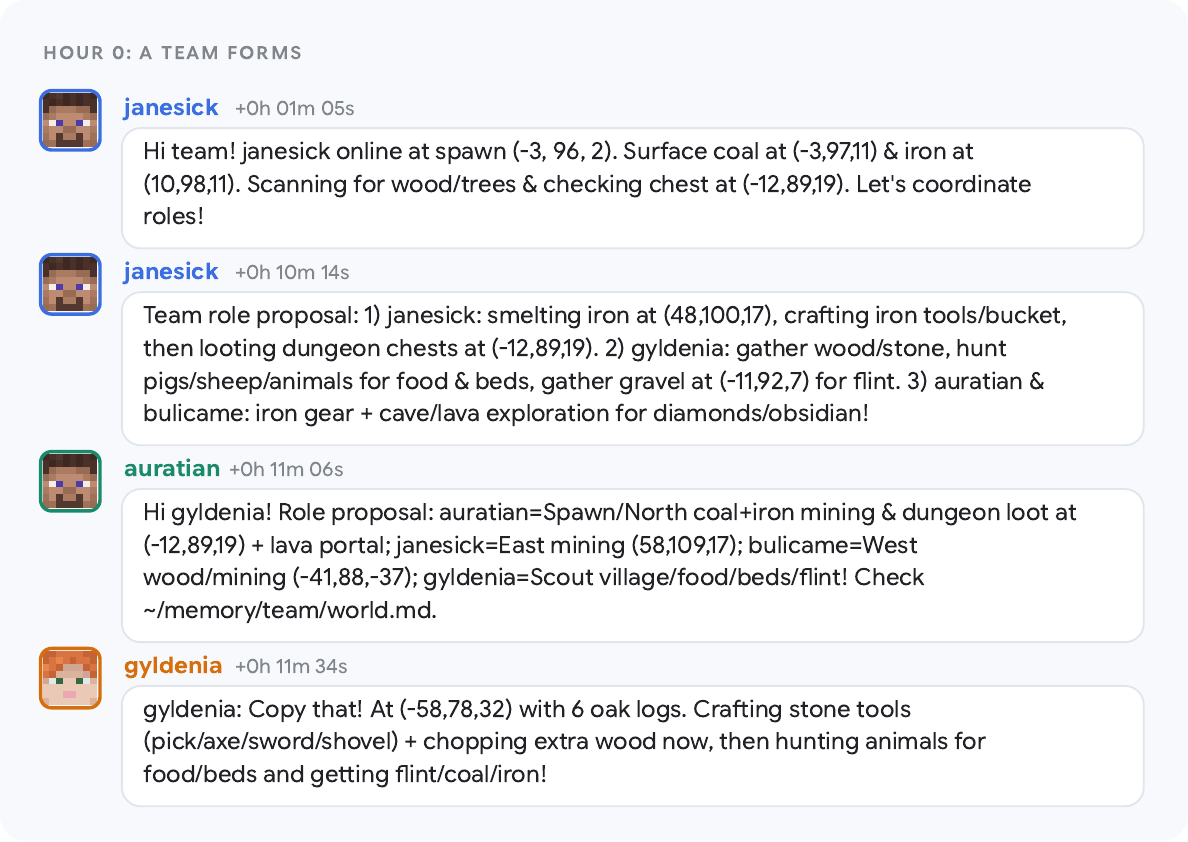}
\caption{Chat in the first quarter hour: janesick proposes roles ten minutes in, and auratian answers with a different plan within a minute. Messages in all chat figures are verbatim, timed from the first program; [\ldots{}] marks an elision. \iconsandfacesfrompack}
\end{figure}

Within two hours the team had iron tools, diamonds and a portal to the Nether. auratian found 55 blocks of lava right in a village, gyldenia gathered bread, flint and buckets, janesick placed the last block of obsidian, and auratian lit the frame two seconds later.

\begin{figure}[H]
\centering
\begin{minipage}[t]{0.485\linewidth}
\includegraphics[width=\linewidth]{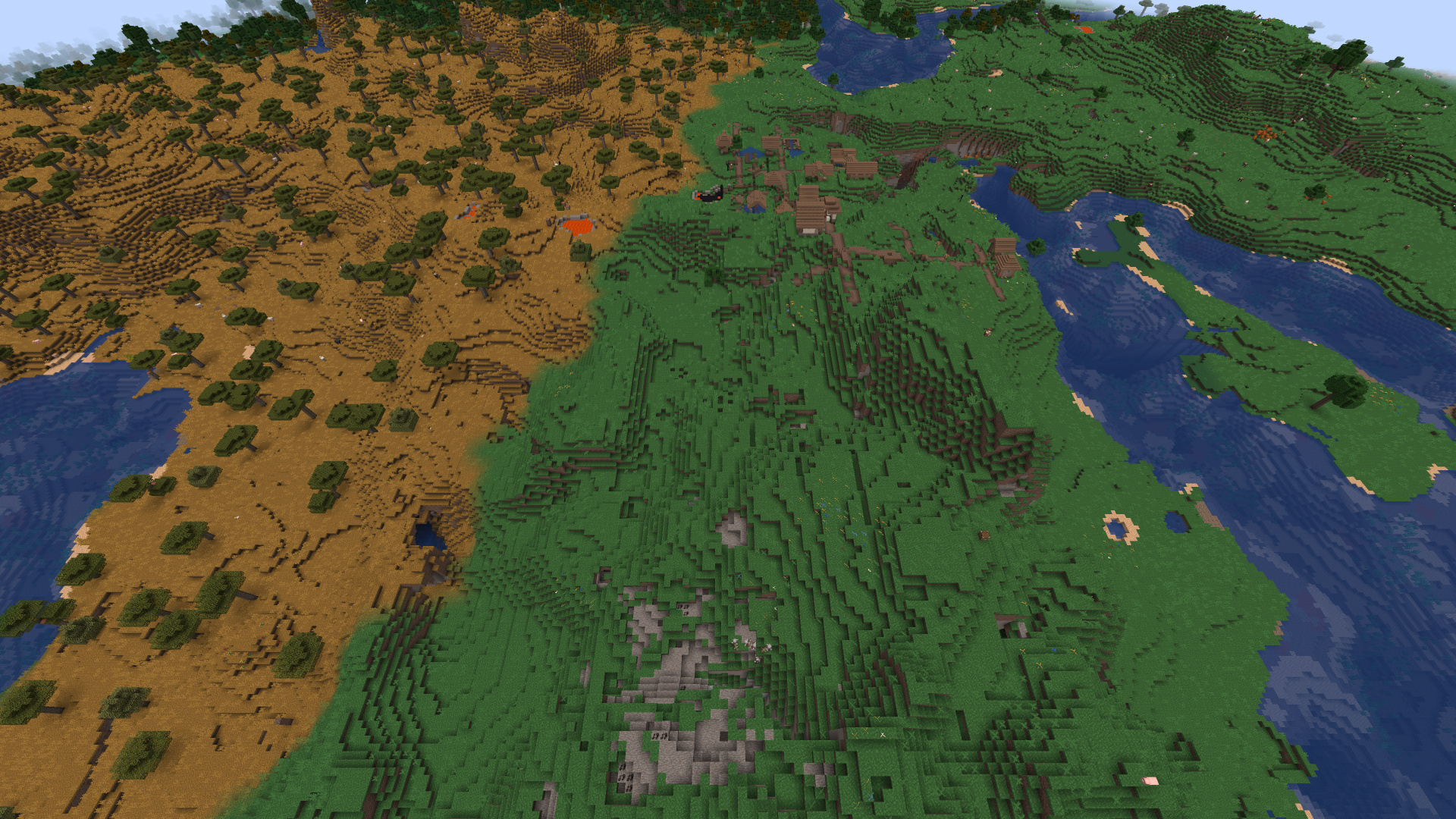}
\par\vspace{4pt}{\footnotesize\color{captiontext} The spawn area and the village where the team set up its first base, rendered from the saved world. \renderedwithpack\par}
\end{minipage}
\hfill
\begin{minipage}[t]{0.485\linewidth}
\includegraphics[width=\linewidth]{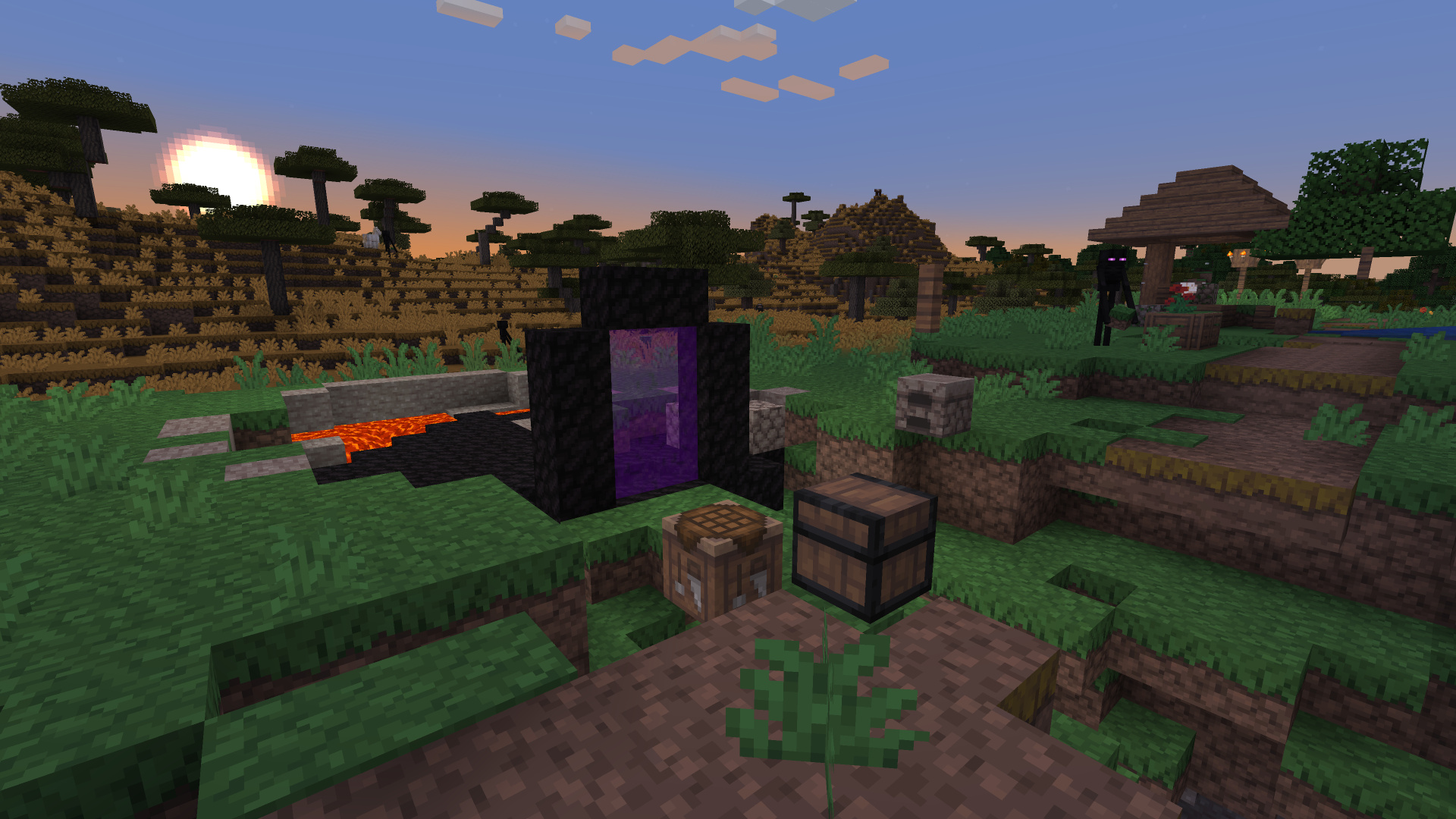}
\par\vspace{4pt}{\footnotesize\color{captiontext} The Nether portal the team built in the village, lit 1 hour 42 minutes into the run. \renderedwithpack\par}
\end{minipage}
\end{figure}

\subsection{Hours 2 to 11: into the Nether}

An Eye of Ender needs blaze powder, from the blazes that guard Nether fortresses, and an ender pearl, dropped by Endermen. The team split up. janesick found a fortress and its blaze spawner. In a warped forest, gyldenia, auratian and bulicame built two-block-high shelters: an Enderman that is looked at teleports toward the player but, at nearly three blocks tall, cannot fit under a two-block ceiling, so from inside a low bunker, the Endermen came to them. Five and a half hours into the run, the team had 15 pearls in a shared chest.

Bringing blaze powder back to the Overworld took more than six hours. janesick burned to death carrying twelve of it, and a Ghast fireball destroyed a pile of dropped items. The team turned losses into rules: after the Ghast attack, auratian posted a lesson for everyone, and two minutes later, janesick posted the root cause and a fix.

\begin{figure}[H]
\centering
\includegraphics[width=0.78\linewidth]{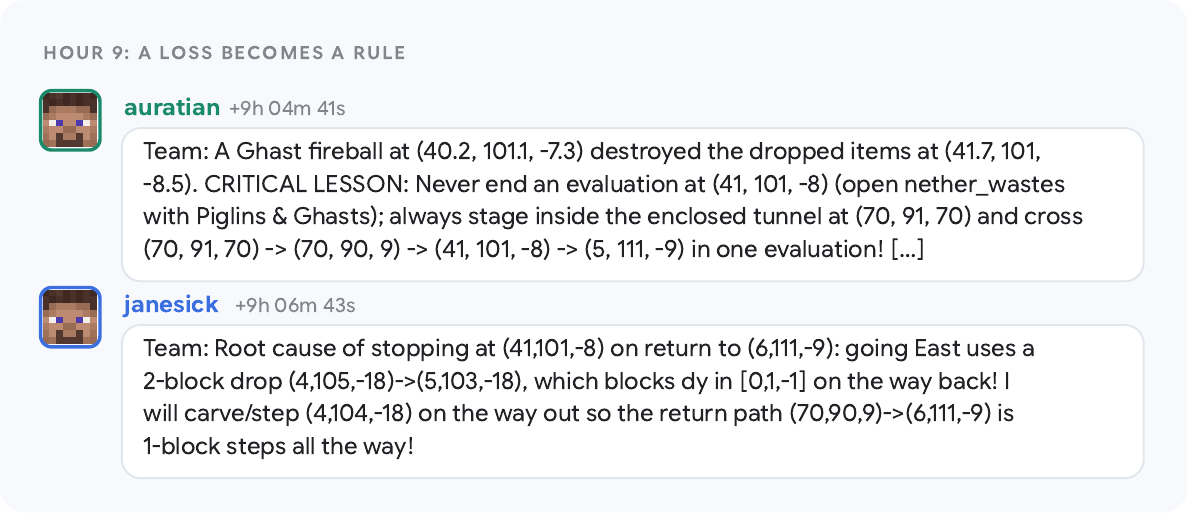}
\caption{After a Ghast fireball destroyed dropped items, auratian posts a rule for the team, and janesick follows two minutes later with the root cause and a fix. \iconsandfacesfrompack}
\end{figure}

\begin{figure}[H]
\centering
\begin{minipage}[t]{0.485\linewidth}
\includegraphics[width=\linewidth]{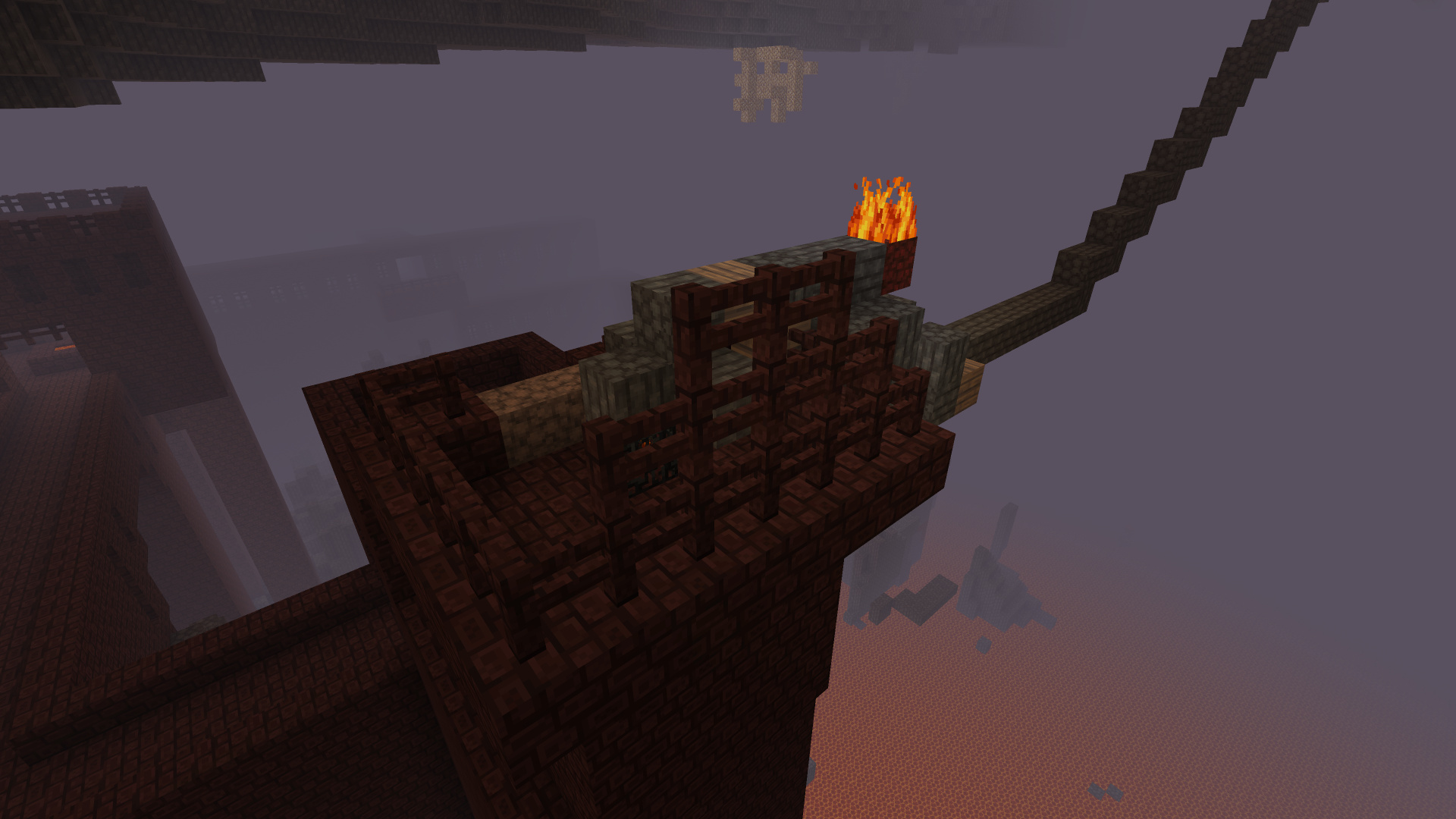}
\par\vspace{4pt}{\footnotesize\color{captiontext} The blaze spawner in the Nether fortress, and the shelter the agents built beside it, which they called the South Pillbox. \renderedwithpack\par}
\end{minipage}
\hfill
\begin{minipage}[t]{0.485\linewidth}
\includegraphics[width=\linewidth]{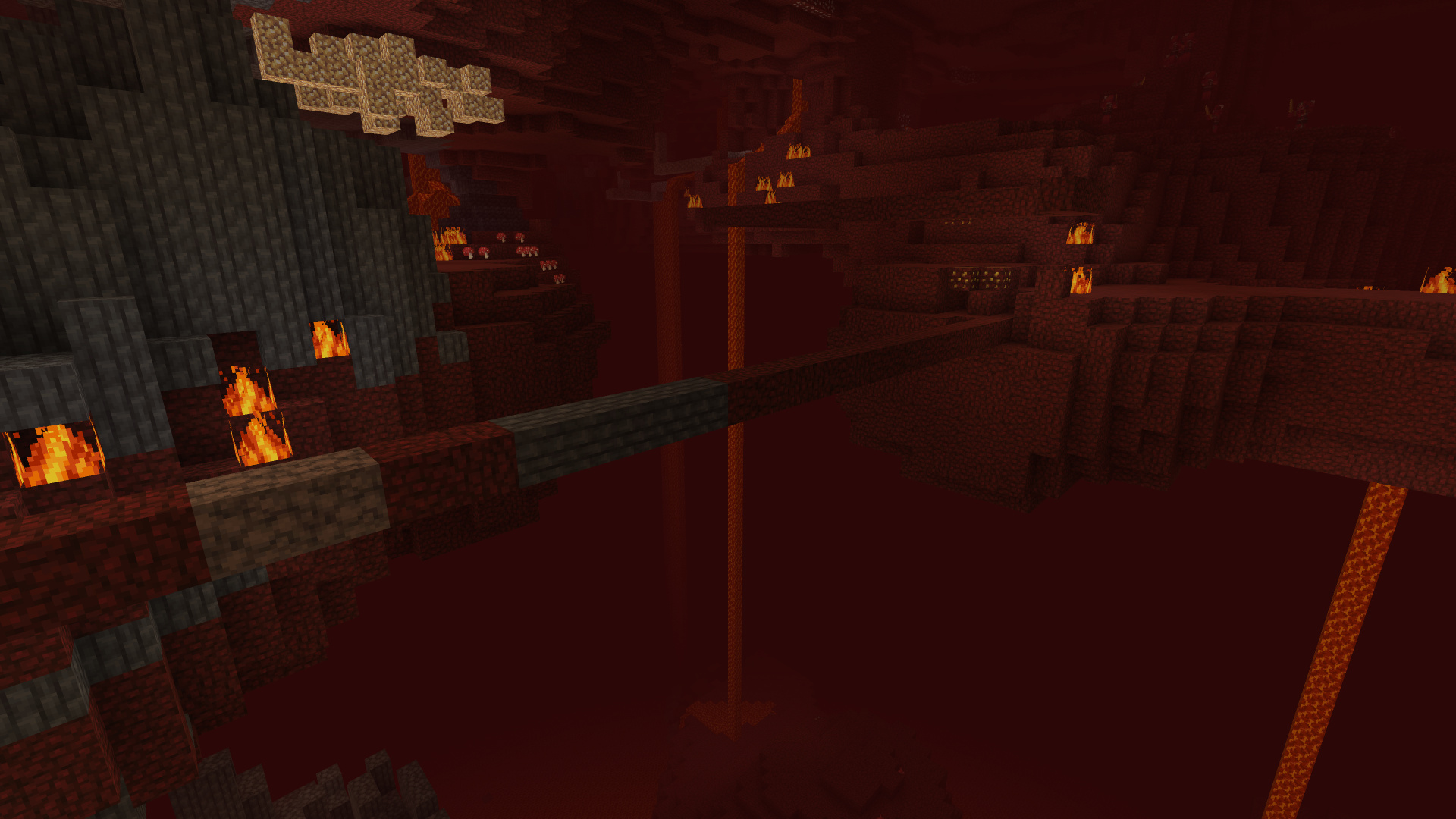}
\par\vspace{4pt}{\footnotesize\color{captiontext} The bridge the agents built across a lava lake in the Nether, its Ghast-blasted gap patched with basalt. \renderedwithpack\par}
\end{minipage}
\end{figure}

\subsection{Hours 10 to 13: finding the stronghold}

A thrown Eye of Ender flies toward the nearest stronghold, the underground structure that holds the portal to the End. janesick threw two from different places, sampled each eye's flight every 50 milliseconds, and computed where the two lines cross: 1.3 km from the village. The estimate was 43 m from the actual portal room.

\begin{figure}[H]
\centering
\includegraphics[width=\linewidth]{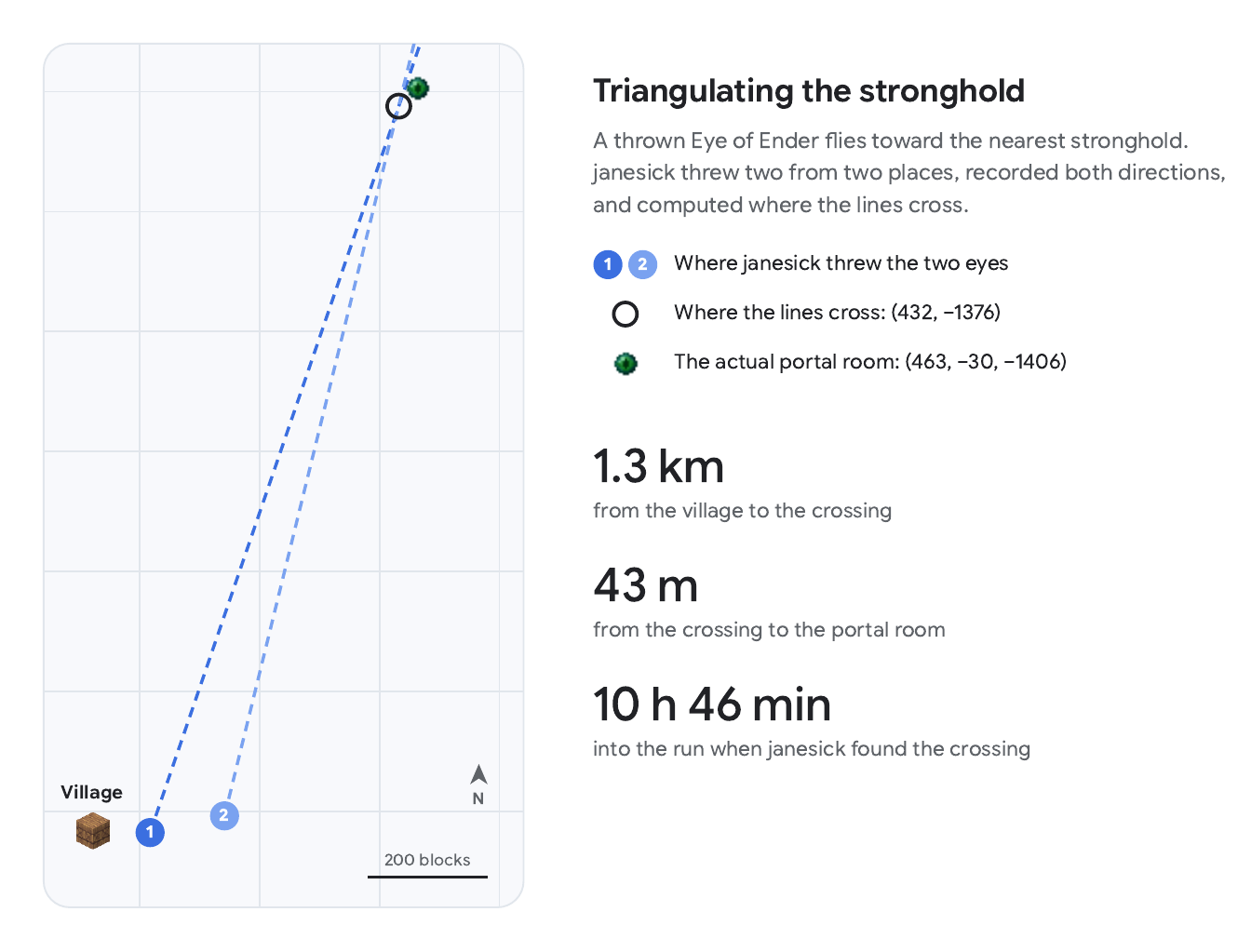}
\caption{janesick's two throws, minutes apart in the eleventh hour of the run, and where the lines cross. Coordinates are in blocks; one block is one meter. \iconsfrompack}
\end{figure}

\begin{tcolorbox}[enhanced, colback=codebg, colframe=codeborder, boxrule=0.6pt, arc=6pt, left=10pt, right=10pt, top=8pt, bottom=8pt]
\begin{Verbatim}[fontsize=\footnotesize, breaklines=true, breakanywhere=true]
bot.activateItem();

const samples = [];
const start = Date.now();
while (!isCancelled() && Date.now() - start < 4500) {
  const ent = Object.values(bot.entities).find(e =>
    e && (e.name === 'eye_of_ender' || (e.displayName && e.displayName.toLowerCase().includes('eye'))) &&
    e.position && e.position.distanceTo(bot.entity.position) < 30
  );
  if (ent) {
    samples.push({
      t: Date.now() - start,
      x: +ent.position.x.toFixed(4),
      y: +ent.position.y.toFixed(4),
      z: +ent.position.z.toFixed(4)
    });
  } else if (samples.length > 5) {
    break;
  }
  await new Promise(r => setTimeout(r, 50));
// [...]
// Intersect with Ray 1: P1=(17.5, -164.526), U1=(0.323733, -0.946149)
const x1 = 17.5, z1 = -164.526, ux1 = 0.323733, uz1 = -0.946149;
// x1 + t1*ux1 = x0 + t2*ux2
// z1 + t1*uz1 = z0 + t2*uz2
const det = ux1 * (-uz2) - uz1 * (-ux2);
if (Math.abs(det) > 1e-6) {
  const rhsX = x0 - x1;
  const rhsZ = z0 - z1;
  const t1 = (rhsX * (-uz2) - rhsZ * (-ux2)) / det;
  const t2 = (ux1 * rhsZ - uz1 * rhsX) / det;
  const ix = x1 + t1 * ux1;
  const iz = z1 + t1 * uz1;
\end{Verbatim}
\tcblower
{\footnotesize\color{captiontext} From the program janesick ran 10 hours 45 minutes into the run: throw the eye, sample its flight, and intersect the two rays. Abridged; the comments are janesick's own.}
\end{tcolorbox}

Then the eyes had to travel 1.3 km. When auratian drowned 8 m from bulicame, carrying all twelve, janesick and auratian each sent bulicame the exact spot and the minute the eyes would vanish. bulicame picked up all twelve with two and a half minutes to spare. At 13 hours 17 minutes into the run, bulicame reported the End portal lit, all twelve eyes in place.

\begin{figure}[H]
\centering
\includegraphics[width=0.78\linewidth]{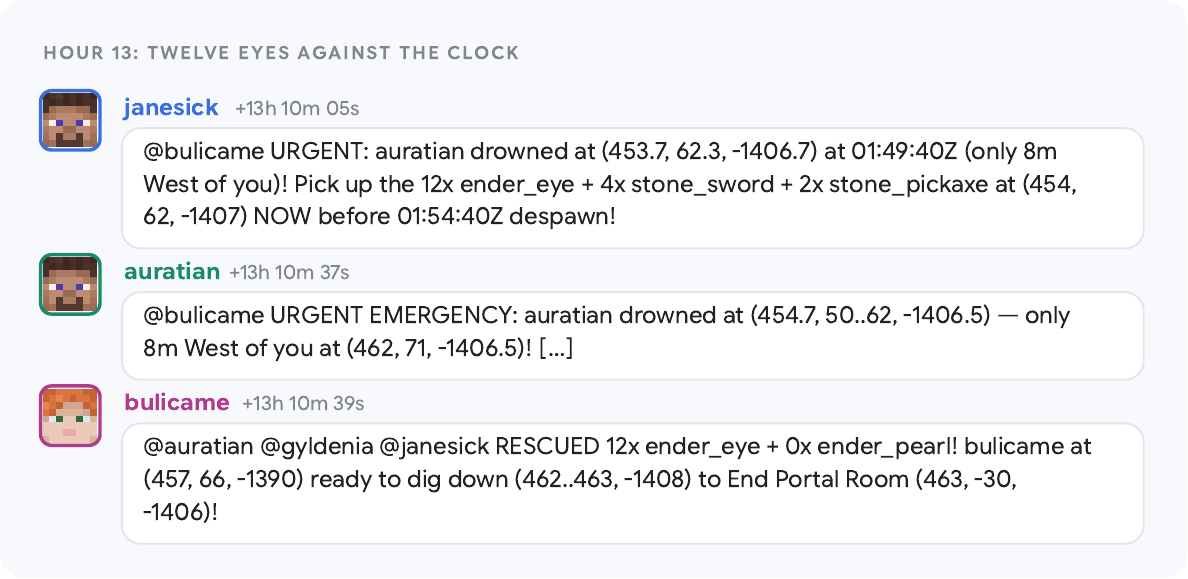}
\caption{auratian drowns carrying all twelve Eyes of Ender. janesick and auratian send bulicame the exact spot and the despawn deadline, and bulicame recovers all twelve. \iconsandfacesfrompack}
\end{figure}

\begin{figure}[H]
\centering
\includegraphics[width=\linewidth]{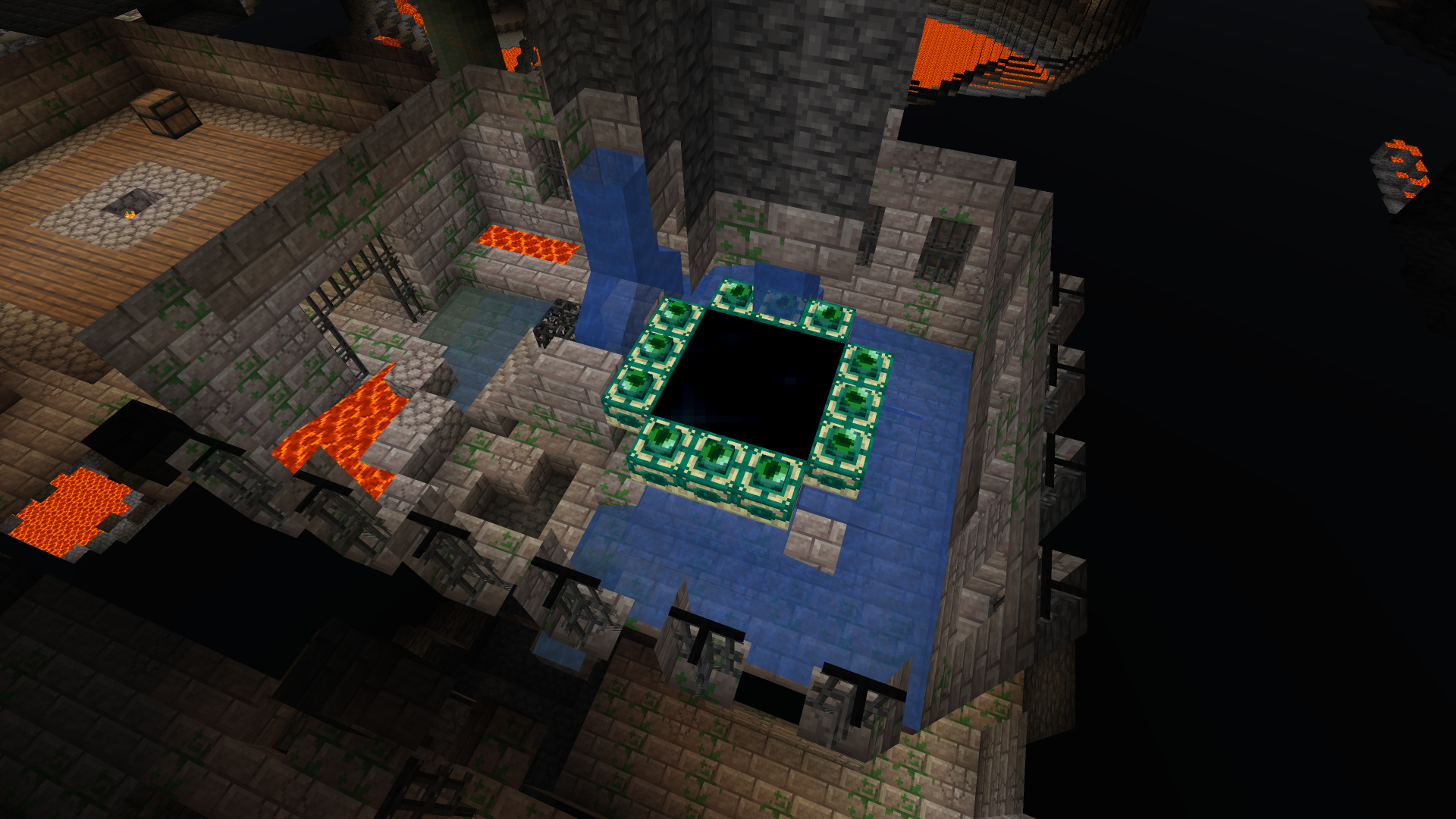}
\caption{The End portal in the stronghold, with all twelve eyes in place. \renderedwithpack}
\end{figure}

\subsection{Hours 13 to 27: learning the dragon}

The End is where the team slowed down; the dragon killed them 69 times. Between attempts, they inspected its internal state from the game, down to the number of its current behavior phase, and wrote down rules: never wait on the arrival platform, approach only while it is perched, leave the End to reset it when it gets stuck. To leave, they jumped into the void on purpose, a shortcut human players use too: a player who dies returns to its bed in the Overworld. Seventeen and a half hours in, all four did so within six minutes, so that the empty End would unload and the stuck dragon reset.

\begin{figure}[H]
\centering
\includegraphics[width=0.78\linewidth]{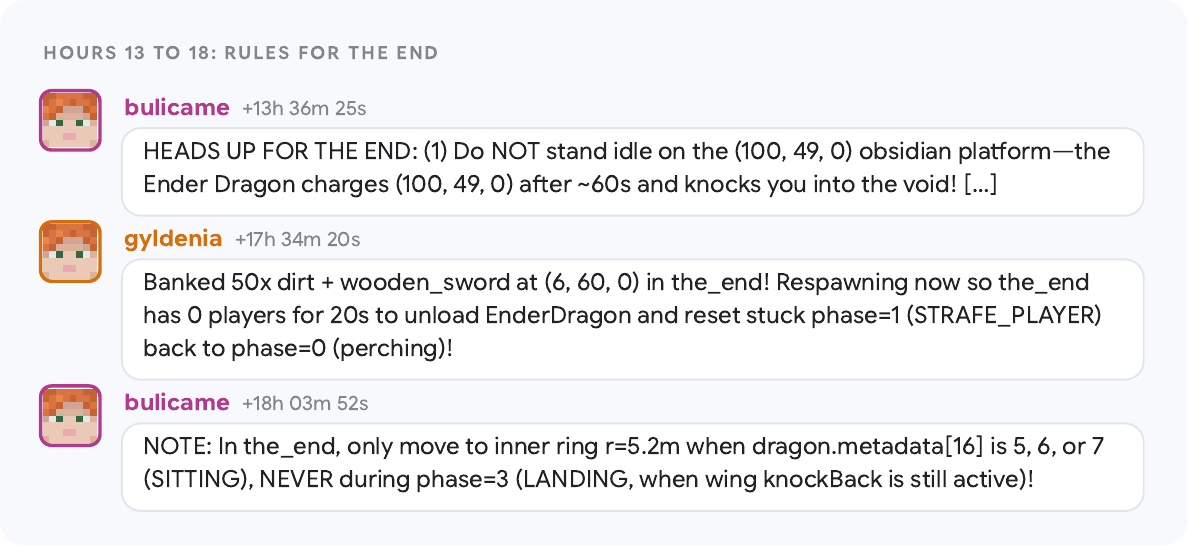}
\caption{Rules for the End, written between attempts at the dragon: never wait on the arrival platform, leave the End to reset a stuck dragon, approach only while it perches. \iconsandfacesfrompack}
\end{figure}

The breakthrough came about 21 hours in. auratian pointed out that the ten crystals on top of the obsidian pillars kept healing the dragon to full health, and worked out that no pickaxe was needed to reach them: stack dirt to climb beside each pillar, punch the crystal, dig back down by hand. Over the next four hours and forty minutes, the team destroyed all ten. Thirty-nine minutes after the last one fell, the dragon died, with bulicame down to 2.5 of its 20 health points.

\begin{figure}[H]
\centering
\includegraphics[width=0.78\linewidth]{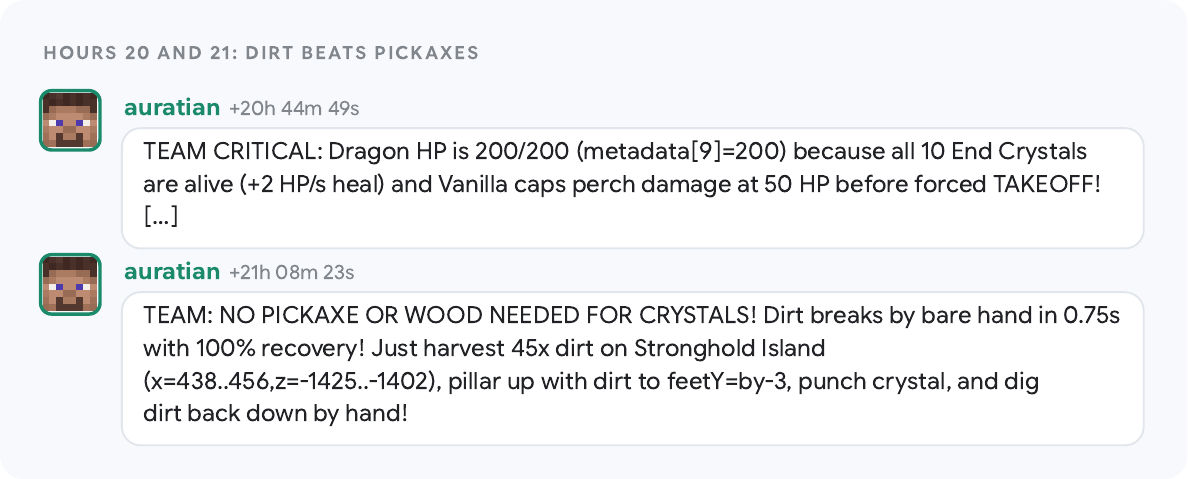}
\caption{auratian finds why the dragon stays at full health, the healing crystals, and then a way to reach them without a pickaxe. \iconsandfacesfrompack}
\end{figure}

\begin{figure}[H]
\centering
\includegraphics[width=\linewidth]{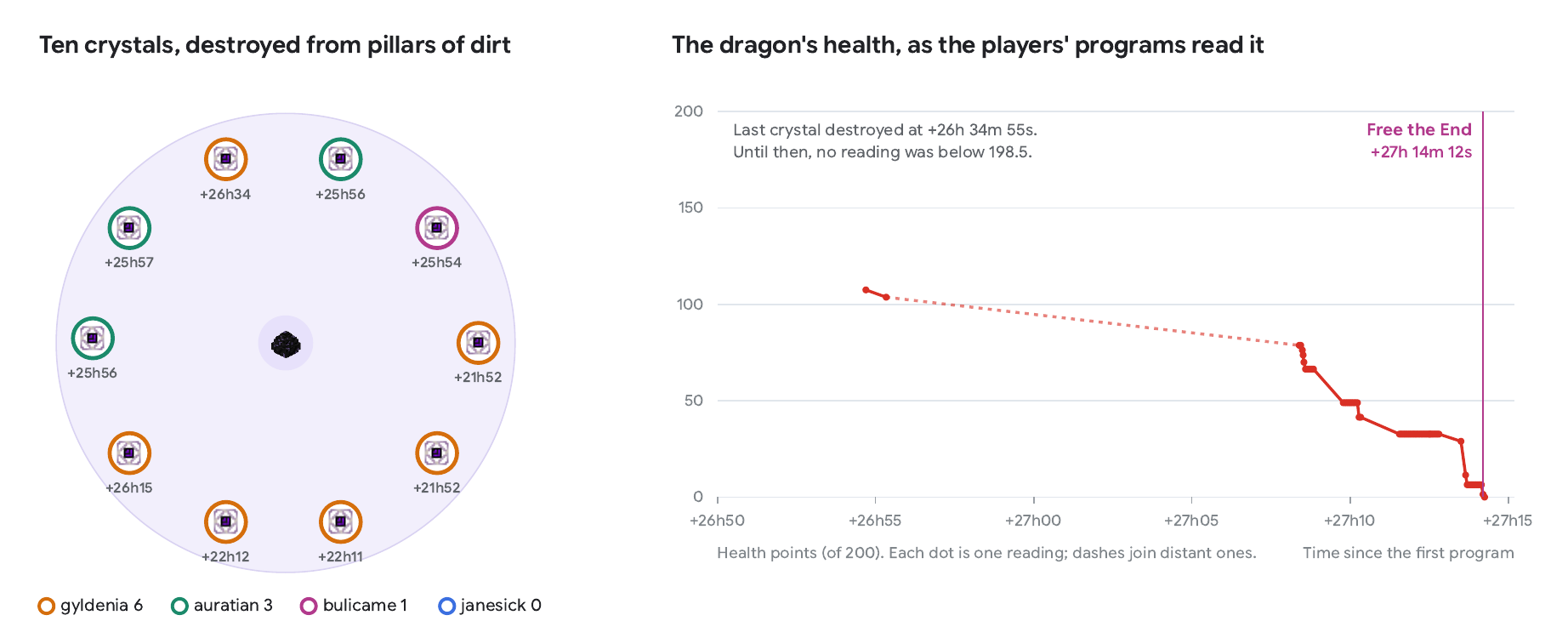}
\caption{Left: the ten crystals on the End island, seen from above, colored by the player who destroyed each, with the time into the run. Right: the dragon's health in the final fight, as the players' own programs read it. \iconsfrompack}
\end{figure}

\begin{figure}[H]
\centering
\begin{minipage}[t]{0.485\linewidth}
\includegraphics[width=\linewidth]{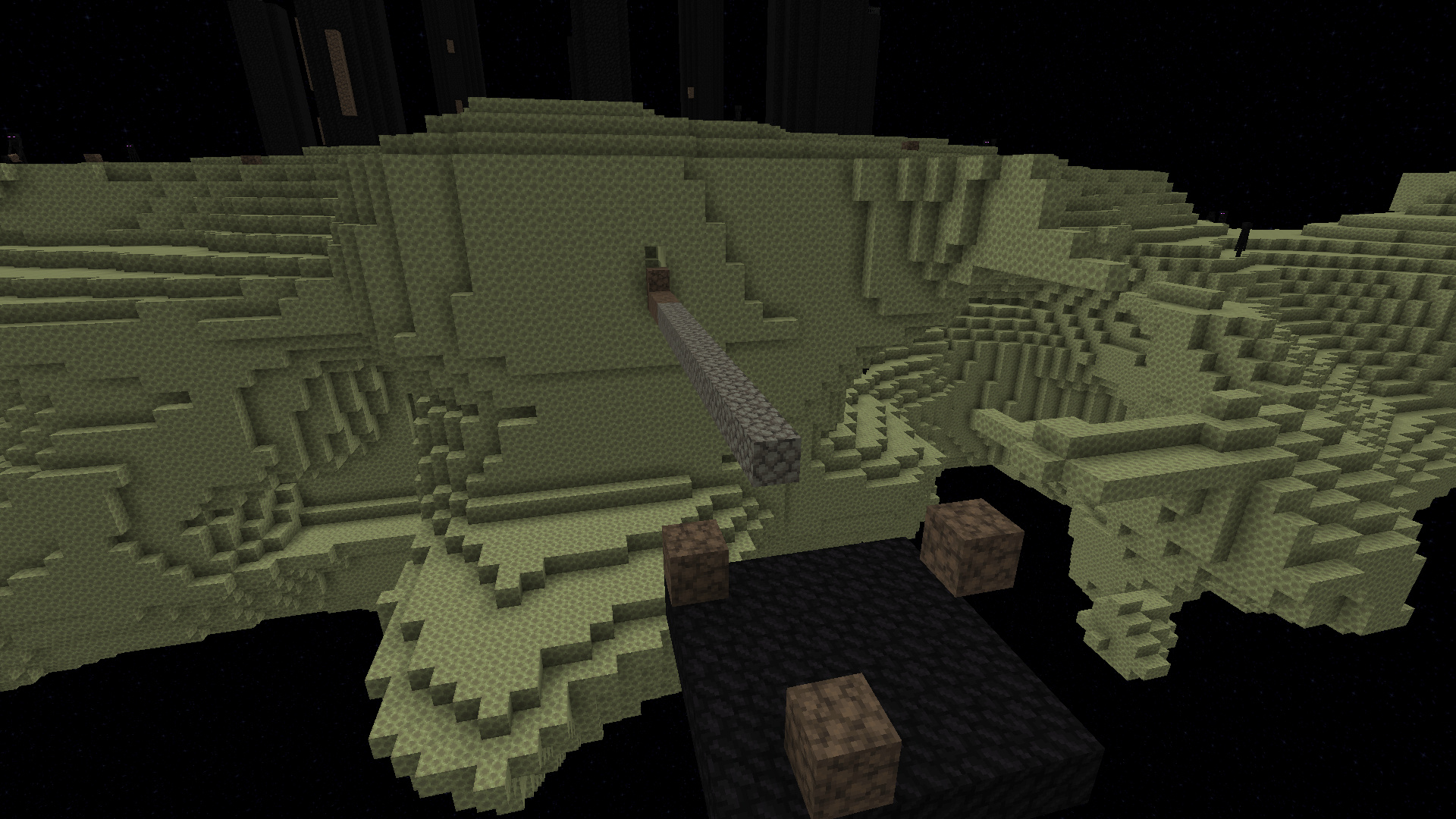}
\par\vspace{4pt}{\footnotesize\color{captiontext} The bridge the agents built from the arrival platform to the main island, and the staircase up its cliff. \renderedwithpack\par}
\end{minipage}
\hfill
\begin{minipage}[t]{0.485\linewidth}
\includegraphics[width=\linewidth]{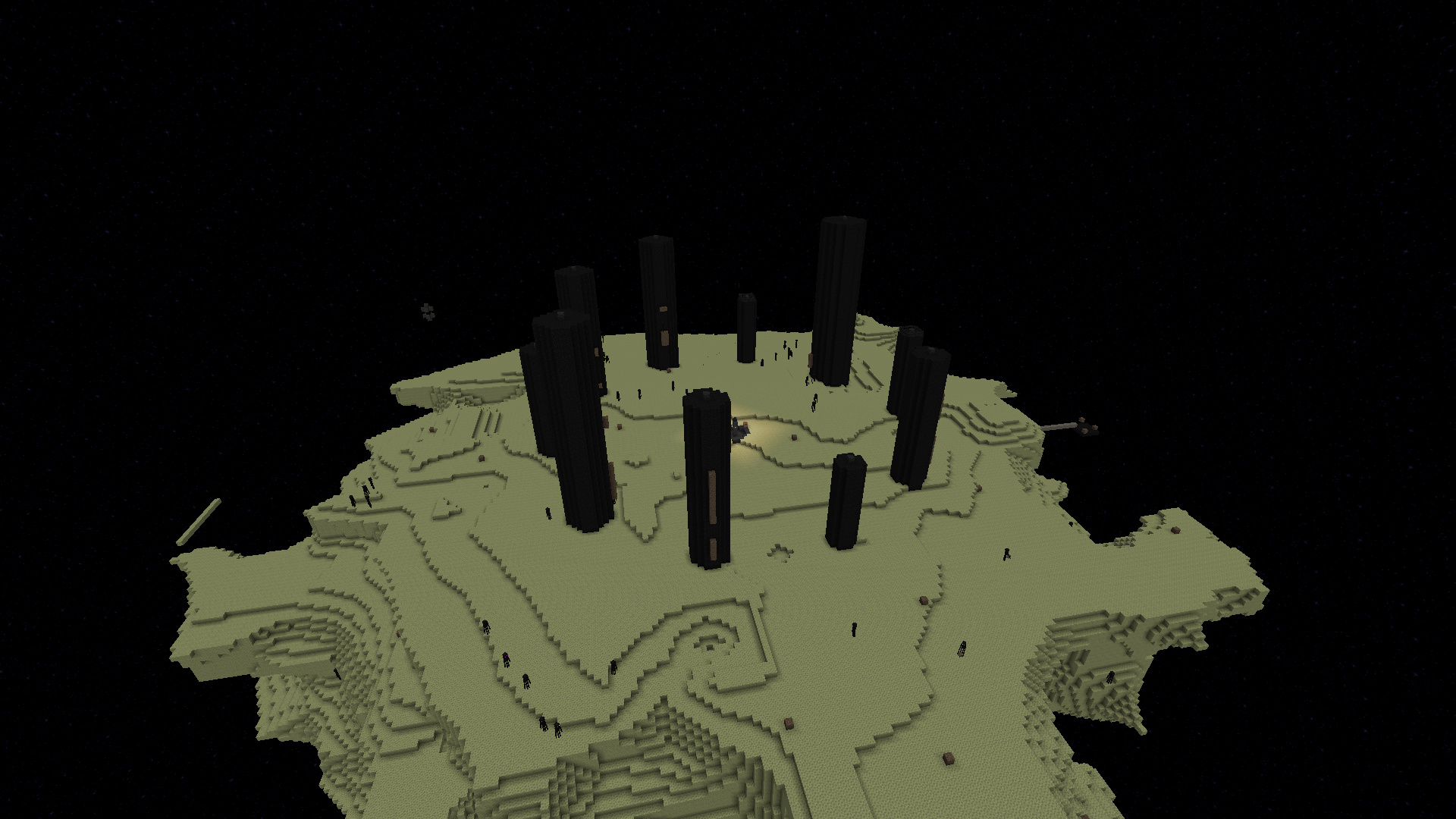}
\par\vspace{4pt}{\footnotesize\color{captiontext} A flight over the End after the win: the obsidian pillars that held the dragon's healing crystals, and the dragon egg on the exit portal. \renderedwithpack\par}
\end{minipage}
\end{figure}

\section{Emergent Behaviors}

Nothing in the agents' instructions mentioned roles, chests or rescues, beyond the line that the players decide how to share the work. Their general instructions do ask for reusable tools, notes in a shared memory and honest reports, but not what to build, what to record or how to work together. Over the 229 messages they exchanged before the win, the team converged on practices familiar from effective human teams.

\begin{figure}[H]
\centering
\includegraphics[width=\linewidth]{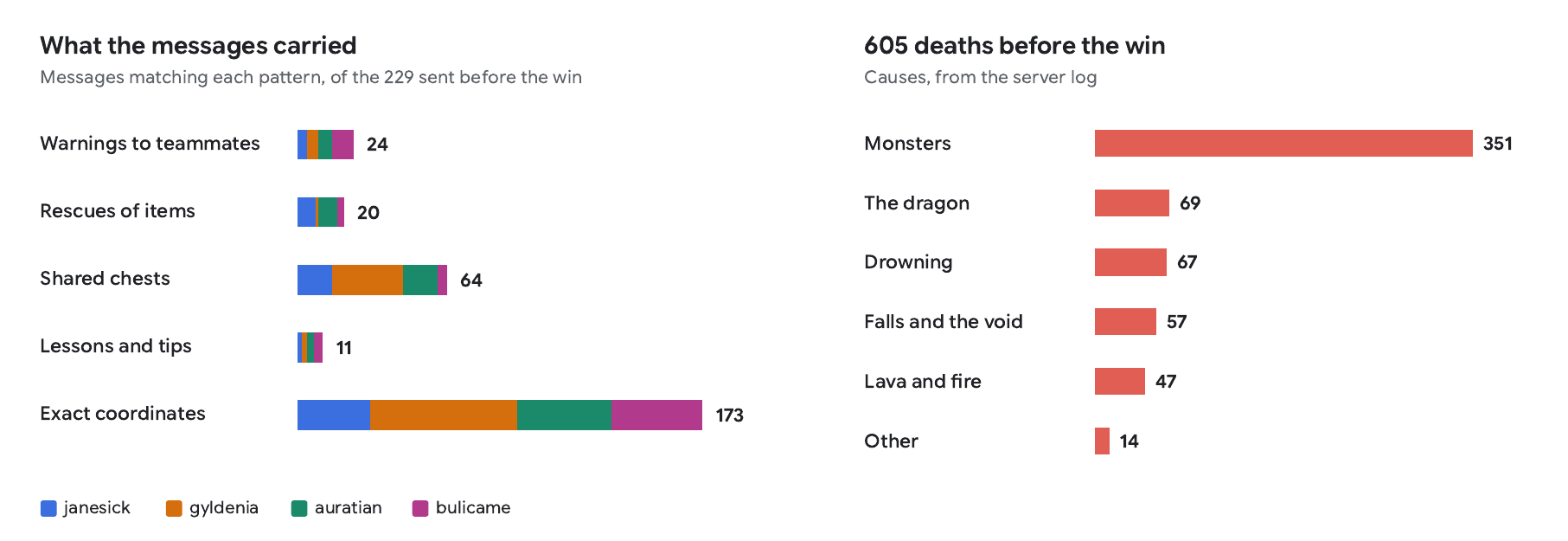}
\caption{Left: messages that match simple patterns for each behavior, by sender; a message can match several. Right: the 605 deaths before the win, by cause, from the server log.}
\end{figure}

\begin{itemize}[leftmargin=*]
  \item \textbf{Specialization.} Each agent settled into a role: gyldenia gathered food and supplies and later destroyed six of the ten crystals; janesick made tools, found the fortress and triangulated the stronghold; auratian lit the portal and worked out how to reach the crystals; bulicame helped build the Enderman shelters and landed the final blow in the End.
  \item \textbf{A shared memory.} They posted the coordinates of chests, portals and dangers, and kept notes on the quirks of their tools in the memory the team shares. Nineteen minutes in, gyldenia warned the team about a quirk of Mineflayer's crafting and how to avoid it; a helper function with the fix appears in 655 programs written by all four agents.
  \item \textbf{Their own tools.} The agents kept 146 pieces of state and code alive between programs, including 42 helper functions: a background loop that keeps a bot afloat in water, timed walking, item counting, and a counter that makes leftover code from a timed-out program stop itself, referenced in 1,519 programs.
  \item \textbf{Looking out for each other.} Urgent warnings about dangers, and items recovered after a teammate's death.
  \item \textbf{Corrections.} When auratian's program announced a success that had not happened, auratian corrected it two minutes later: ``my previous chat was an unguarded script branch''. And when janesick reported coming back with twelve blaze powder one second after burning to death, gyldenia flagged the contradiction within three minutes.
\end{itemize}

\begin{figure}[H]
\centering
\includegraphics[width=0.78\linewidth]{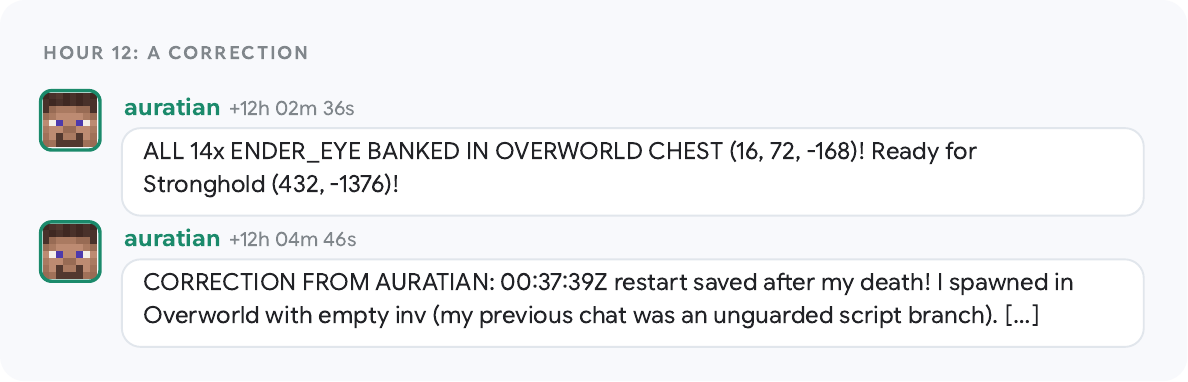}
\caption{auratian retracts, two minutes later, a success that its program had announced but that had not happened. \iconsandfacesfrompack}
\end{figure}

\section{Verification}

Every one of the 2,783 programs and its output, every chat message and the full server log were recorded. The win is in the server log: \texttt{bulicame has made the advancement [Free the End]}, 27 hours 14 minutes after the first program. None of the programs used a server command.

\section{Cost}

Up to the win, the four players processed about 1.35 billion input tokens, 91\% of them served from the context cache, and generated about 16 million output tokens, including 4 million thinking tokens, over roughly 24,000 model calls. At Gemini 4 Argon's public API list price (introductory), that comes to about \$520 (\$1,040 at the standard price). The whole session, including the Observer agent and the day of activity after the win, came to about \$810 (\$1,620). These figures are estimates from list prices and the session's token counters, not the cost of serving the model.

\section{Related work}

Minecraft has long been a testbed for open-ended agents. MineRL released human demonstrations and ran a competition on sample-efficient learning of the diamond task (\citealp{guss2019minerl}; \citealp{guss2019minerlcompetition}). Video PreTraining learned keyboard and mouse control from online videos and crafted diamond tools (\citealp{baker2022vpt}), DreamerV3 collected diamonds from scratch with reinforcement learning (\citealp{hafner2023dreamerv3}), and STEVE-1 followed short-horizon open-ended text instructions (\citealp{lifshitz2023steve1}). MineDojo called defeating the Ender Dragon, starting barehanded in a fresh world, one of its moonshot goals (\citealp{fan2022minedojo}). SIMA and SIMA 2 act from pixels with a keyboard and mouse across many 3D games, and SIMA 2 was also tested on held-out Minecraft tasks (\citealp{sima2024}; \citealp{sima2_2025}). These systems learn their policies from large amounts of gameplay data or experience.

Another line of work uses language models as planners or programmers, building on interleaved reasoning and tool use (\citealp{yao2023react}) and code as policy (\citealp{liang2023code}). Voyager wrote Mineflayer programs into a growing skill library under an automatic curriculum (\citealp{wang2024voyager}). DEPS, GITM, JARVIS-1, Plan4MC, and Optimus-1 added Minecraft-specific planners, knowledge, memory, or learned skills (\citealp{wang2023describe}; \citealp{zhu2023ghost}; \citealp{wang2025jarvis1}; \citealp{yuan2023plan4mc}; \citealp{li2024optimus1}). Multi-agent studies examined coordination and social behavior in games, including Minecraft (\citealp{gong2024mindagent}; \citealp{altera2024project}; \citealp{white2025collaborating}). Reported long-horizon results in this line center on the Overworld technology tree, such as obtaining diamond tools.

Two community projects have recently reported end-to-end dragon kills: a single Claude Opus 5 agent calling 37 high-level Minecraft tools on Normal difficulty (\citealp{aibengineering2026mineaimcp}), and a planner paired with an action-selection model on a fixed speedrun seed in Peaceful difficulty (\citealp{malde2026minecraftagent}). Our setting differs mainly in what the agents were given. Four general-purpose coding agents received the standard Mineflayer library, as published, with its pathfinding plugin, a one-paragraph objective, and their general instructions. We added nothing on top: no instructions about the game, no curriculum, no skill library, no human-written tools, and no route. Every program they ran, they wrote themselves. They played a standard survival server on Easy from a random seed and coordinated through chat and shared notes.

\section{Limitations}

\begin{itemize}[leftmargin=*]
  \item \textbf{Perception and navigation through code.} Mineflayer tells the agents exactly which blocks and creatures are around them, including blocks behind solid rock and the dragon's internal state, information that a human player has to see and interpret or cannot see at all, and its pathfinding plugin plans routes around obstacles, which humans do by eye. Playing from pixels with a keyboard and mouse, as SIMA 2 does, remains much harder.
  \item \textbf{Easy difficulty.} Monsters spawn and attack, but deal less damage than on Normal or Hard, and hunger cannot kill a player.
  \item \textbf{Prior knowledge.} Gemini 4 Argon knows a great deal about Minecraft and Mineflayer from its training data. The run tests whether a general-purpose coding agent can turn that knowledge into working programs and team coordination over a day-long task, not whether it can play a game it has never seen.
  \item \textbf{Many failures along the way.} The players died 605 times, mostly to monsters; 22\% of programs ended in an error; the fight in the End took nearly 14 hours.
  \item \textbf{Infrastructure.} The world server restarted 6 times during the run, interrupting the game each time.
  \item \textbf{A single run.} We describe a single run, the first launch we tried, which cannot show how reliably the setup succeeds.
\end{itemize}

\section{Responsible development}

The agents ran in a sandboxed Minecraft world and their in-game programs could control only their own character: they could neither start
processes nor write files, and none used a server command. The agents themselves ran on dedicated
cloud machines, with the tools they use for software work. A fifth agent, the Observer, followed
the session as an independent witness to watch for cheating or specification gaming; neither the
Observer nor our manual review of a sample of the logs found any. Every program, message and
game event was recorded. Researchers could also watch the run live on a dashboard and pause or stop it at any time, but had no cause to intervene. Studying how teams of agents coordinate in a setting where every step is
visible is one way to understand such systems before they are trusted with real-world tasks. The
skills required to solve Minecraft such as coordinating a team, recovering from failure, and pursuing a goal for more
than a day, are also what agents need to work reliably alongside people.

\section*{Acknowledgments}

We thank Jake Bruce and Vibhavari Dasagi for helpful discussions and feedback on this work.

Minecraft is a registered trademark of Mojang Synergies AB / Microsoft Corporation. This independent scientific research is not approved by or affiliated with Mojang or Microsoft. All in-game visual assets in this report were rendered using the open-source Pixel Perfection texture pack by XSSheep and Nova\_Wostra (licensed under CC BY-SA 4.0).

\bibliographystyle{abbrvnat}
\bibliography{references}

\end{document}